\documentclass{article}

\usepackage{arxiv}

\usepackage[utf8]{inputenc} 
\usepackage[T1]{fontenc}    
\usepackage{hyperref}       
\usepackage{url}            
\usepackage{booktabs}       
\usepackage{amsfonts}       
\usepackage{nicefrac}       
\usepackage{microtype}      
\usepackage{lipsum}
\usepackage{graphicx}

\usepackage{multirow}%
\usepackage{amsmath,amssymb}%
\usepackage{amsthm}%
\usepackage{mathrsfs}%
\usepackage[title]{appendix}%
\usepackage{xcolor}%
\usepackage{textcomp}%
\usepackage{manyfoot}%
\usepackage{algorithm}%
\usepackage{algorithmicx}%
\usepackage{algpseudocode}%
\usepackage{listings}%
\usepackage{subcaption}

\title{What Differentiates Educational Literature? A Multimodal Fusion Approach of Transformers and Computational Linguistics}

\author{
 Jordan J. Bird \\
  Department of Computer Science\\
  Nottingham Trent University\\
  Nottingham, Nottinghamshire, NG11 8NS, United Kingdom \\
  \texttt{jordan.bird@ntu.ac.uk} \\
}

\begin{document}
\maketitle
\begin{abstract}
The integration of new literature into the English curriculum remains a challenge since educators often lack scalable tools to rapidly evaluate readability and adapt texts for diverse classroom needs. This study proposes to address this gap through a multimodal approach that combines transformer-based text classification with linguistic feature analysis to align texts with UK Key Stages. Eight state-of-the-art Transformers were fine-tuned on segmented text data, with BERT achieving the highest unimodal F1 score of 0.75. In parallel, 500 deep neural network topologies were searched for the classification of linguistic characteristics, achieving an F1 score of 0.392. The fusion of these modalities shows a significant improvement, with every multimodal approach outperforming all unimodal models. In particular, the ELECTRA Transformer fused with the neural network achieved an F1 score of 0.996. Unimodal and multimodal approaches are shown to have statistically significant differences in all validation metrics (accuracy, precision, recall, F1 score) except for inference time. The proposed approach is finally encapsulated in a stakeholder-facing web application, providing non-technical stakeholder access to real-time insights on text complexity, reading difficulty, curriculum alignment, and recommendations for learning age range. The application empowers data-driven decision making and reduces manual workload by integrating AI-based recommendations into lesson planning for English literature.
\end{abstract}

\section{Introduction}
The integration of new literature into education remains a significant challenge for educators, who often lack access to robust tools to evaluate and adapt texts for classroom settings. These issues are further exacerbated by the need to respond to trends and integrate popular contemporary works to retain student interest and enhance learning experiences. Currently, there are no scalable solutions that enable educators to respond quickly to trends by autonomously analysing the complexity of the text, aligning the literature with the appropriate educational stages, and generation of actionable insights for use in the education system. 

This lack of tools thus leaves educators dependent on manual evaluation, which is a resource-intensive process at a time where education systems face ongoing issues of growing class sizes, budget cuts, and work-related stress leading to poor retention. In addition to these problems, decentralised manual evaluation can also lead to inconsistencies in capturing the nuanced demands of a diverse classroom. 

Popular books are wide-ranging in their complexities, thematic depths, and linguistic sophistication, which can make it difficult to determine their suitability across different educational stages. For example, Harper Lee's \textit{To Kill a Mockingbird}, a text commonly found in the classroom, presents distinct challenges in aligning with specific educational stages. The book is written with relative accessibility and narrative style, making it appropriate for students in Key Stage 3 to develop their comprehension skills. In addition, the work explores themes such as growing up, the loss of innocence, and moral development, which are themes that resonate with the early stages of secondary education. In upper Key Stage 3, more complex discussions on systemic racism, class structure, and justice require a higher level of maturity and critical thinking. Beyond, towards Key Stages 4 and 5, Lee's work offers deeper opportunities to analyse thematic complexities and rhetorical devices in social contexts. These examples show the importance of analysis tools for identification of texts that are appropriate for a given learner. 

When a new book is published that becomes significantly popular, or \textit{goes viral}, with young audiences, there exists no analysis at this level yet. If work is to be integrated into the education system, analysis must first be performed to discover which learners the literature is most useful for, such as cross-referencing with the national curriculum. For example, the work may contain useful examples of compound-complex sentences and utilise sophisticated punctuation, making it a useful work to assist in Key Stage 4 education. The issue here lies in the need for a rapid response, that is, by the time manual analysis is exhaustively performed and communicated, young audiences may have moved on to the next popular work. Thus, the insights may no longer remain actionable, and therefore expert time has been wasted. The education system is increasingly shifting towards data-driven teaching approaches that enable more granular personalisation and recommendations. Hence, there is a critical need for innovative approaches that empower educators to make informed decisions while reducing their workload. The work in this article proposes to address this knowledge gap through the use of Natural Language Processing (NLP), computational linguistics, and Artificial Intelligence (AI) to research and develop a stakeholder-facing toolkit for literature analysis. The tools produced following the experiments performed in this study enable teachers to respond proactively to popular and emerging texts, which could enhance curricula with data-driven decisions. 

The main scientific contributions of this work are threefold. Firstly, the studies in this article introduce a novel method for the combination of transformer-based text classification and linguistic feature analysis. The results show that the multimodal approach is far superior to unimodal models in detecting the appropriateness of educational stage of a given literature. Second, the findings of these studies are encapsulated within a web-based toolkit for stakeholders to analyse and visualise text complexity, reading levels, and vocabulary importance, supporting data-driven curriculum development within the field of AI in Education. Finally, the dataset generated for the purposes of this study is publicly released for interdisciplinary analysis and research by the academic community. 

The remainder of this article is as follows. Section \ref{sec:relatedwork} explores notable related work in the field. Section \ref{sec:method} then describes the method followed in the experiments for data collection, machine learning, computational linguistics, and stakeholder-facing web application development. The results of the single and multimodal experiments are presented and contrasted in Section \ref{sec:resultsdiscussion}, before conclusions are drawn and future work arising from this study is proposed in Section \ref{sec:conclusion}.

\section{Related Work}
\label{sec:relatedwork}
The use of AI within educational processes has been observed to promote the personalisation of teaching materials, improve lesson planning procedures, promote efficiency, and create novel experiences to inspire students \cite{chen2020artificial,zhai2021review,holmes2022state}. From a learner's perspective, exciting new methods of learning can be experienced and social inequalities can be alleviated through personalisation of the learning experience. For educators, technological assistance can alleviate workload demands, helping to maintain teaching quality while positively impacting both physical and mental well-being.

Readability assessment is a particularly difficult task that forms an important open issue in the field. Zamanian and Heydari \cite{zamanian2012readability} provide a background of readability formulae and their reliance on features such as sentence length, word length, and frequencies. While these metrics held potential in early research, they are increasingly critically analysed and often fail to account for deeper semantic and domain-level features within a text. The authors note that scores from Flesch and Dale-Chall, for example, can provide estimates of reading difficulty, they cannot measure more intelligent concepts such as audience engagement. Adding to this discourse, the issues were outlined in depth by a letter to the editor from Alzaid, Ali, and Stapleton\cite{alzaid2024limitations}, who critically analyse traditional readability metrics and note that they focus predominantly on quantifiable properties, for example, readability scores, presence of part-of speech, diversity metrics, richness metrics, etc. which do not encapsulate qualitative features such as domain-level features. There also exist inconsistencies between formulae such as Flesch-Kincaid and SMOG, which are two of several features used in the unimodal linguistics model approach in this study prior to multimodal studies. 

In \cite{sung2015constructing}, Sung et al. note the difficulty in autonomously recognising readability, with traditional methods often resulting in low classification accuracy. In the study, the authors proposed the use of linguistic features of four categories (word, semantics, syntax, and cohesion) for the analysis of Chinese text. The results showed that multilevel Support Vector Machine models could achieve 71.75\% classification accuracy. Similarly, \cite{martinc2021supervised} also note that traditional linguistic features often do not allow machine learning models to generalise. The authors propose a deep learning-driven Ranked Sentence Readability Score, which is noted to correlate with human-assigned readability scores. In particular, BERT-based models are noted to outperform temporal and hierarchy-based models across English and Slovenian texts. The authors note the need for domain-specific challenges, which is the focus of this work. Lee et al. \cite{lee2021pushing} build on these findings, arguing that the fusion of text classifiers and transformers achieve state of the art performance, achieving around 99\% classification accuracy on the OneStopEnglish dataset. The authors highlight a complementary relationship between traditional features and transformers, suggesting that multimodality is a potential solution to open issues in the area. 

Open issues relating to traditional readability formulae are also emphasised by Crossley et al.\cite{crossley2023using}, who note that text features alone often do not generalise across specific contexts. The authors propose new readability formulae using machine learning approaches, including BERT. In relation to this study, the proposed approaches were observed to be practical in terms of ability and computational complexity, making them more likely to be appropriate for use on consumer-level hardware in schools. 

In 2021, Ehara proposed the LURAT readability assessment toolkit\cite{ehara2021lurat}, designed for second-language learners. The proposed approach focused on vocabulary tests to estimate the difficulty of learning a word, allowing a learner-centric assessment with consideration of second language learner knowledge. The results showed that LURAT could outperform large language models while being considerably less computationally complex. Authors explored the language learning problem in \cite{uccar2024exploring}, where various machine learning models were trained to assess the readability of multilingual scientific documents. The authors noted high accuracy and F1 scores during training, but noted issues with generalisation to non-training data, observing a drop from 87.33\% to around 34\% to 36\% accuracy on unseen data. 

Literature review has shown that traditional readability research and metrics provide a useful foundational framework; however, they fail to encapsulate the complexity of educational texts by issues such as a lack of semantic depth and inability to handle complex nuances within a text. Recent advances have shown the potential of educator and learner-centred approaches considered within models. This need for additional nuance through extension of traditional approaches leads to the potential of multimodality, where it becomes possible to consider the fusion of these approaches alongside text-based deep learning approaches to alleviate open issues in the field. In addition, current state-of-the-art methods of Large Language Models such as ChatGPT, LLaMa, and Mistral etc. are considerably computationally complex in comparison with the hardware accessible by educators, making accessibility difficult. This study builds on the current state of the art by proposing a multimodal framework which fuses the aforementioned traditional approaches with text-based deep learning approaches, aiming to increase classification performance while maintaining model complexity, selecting pareto-optimal models in the trade-off between capability and real-world accessibility.

\section{Method}
\label{sec:method}
This section contains an overview of the methodology followed by this work from data collection to unimodal model training, to subsequent multimodal model training and comparison. In addition, this section also describes the technical design of the stakeholder-facing web application.

\subsection{Data Collection and Preprocessing}
\label{subsec:datacollection}

\begin{table}[]
\centering
\caption{Project Gutenberg bookshelves and collections used for data collection. }
\label{tab:bookshelves}
\footnotesize
\begin{tabular}{@{}lll@{}}
\toprule
\textbf{Bookshelf}                    & \textbf{Collection}                 & \textbf{Gutenberg Collection ID} \\ \midrule
\multirow{6}{*}{Children's Bookshelf} & Children's Literature               & 20                               \\
                                      & Children's Book Series              & 17                               \\
                                      & Children's Fiction                  & 18                               \\
                                      & Children's Christmas                & 23                               \\
                                      & Children's Myths, Fairy tales, etc. & 216                              \\
                                      & Children's Anthologies              & 213                              \\
\multirow{7}{*}{Fiction}              & Science Fiction                     & 68                               \\
                                      & Gothic Fiction                      & 39                               \\
                                      & Horror                              & 42                               \\
                                      & Adventure                           & 82                               \\
                                      & Detective Fiction                   & 30                               \\
                                      & Fantasy                             & 36                               \\
                                      & Western                             & 77                               \\
\multirow{2}{*}{Classics}             & Harvard Classics                    & 40                               \\
                                      & Classic Antiquity                   & 24                               \\
\multirow{6}{*}{Uncategorised}        & Culture/Civilization/Society        & 432                              \\
                                      & Literature                          & 458                              \\
                                      & Fiction                             & 486                              \\
                                      & Movie Books                         & 49                               \\
                                      & Precursors of Science Fiction       & 62                               \\
                                      & Children \& Young Adult Reading     & 429                              \\ \bottomrule
\end{tabular}
\end{table}

\begin{table}[]
\centering
\caption{UK Key Stage equivalences to Lexile Scores.}
\label{tab:lexile}
\footnotesize
\begin{tabular}{@{}ll@{}}
\toprule
\textbf{Lexile Score} & \textbf{Label} \\ \midrule
\textless 400         & KS1            \\
400 to 800            & KS2            \\
801 to 1000           & KS3            \\
1001 to 1200          & KS4            \\
\textgreater 1200     & KS5            \\ \bottomrule
\end{tabular}
\end{table}

\begin{figure}[]
    \centering
    \begin{subfigure}[b]{0.48\textwidth}
        \centering
        \includegraphics[width=\textwidth]{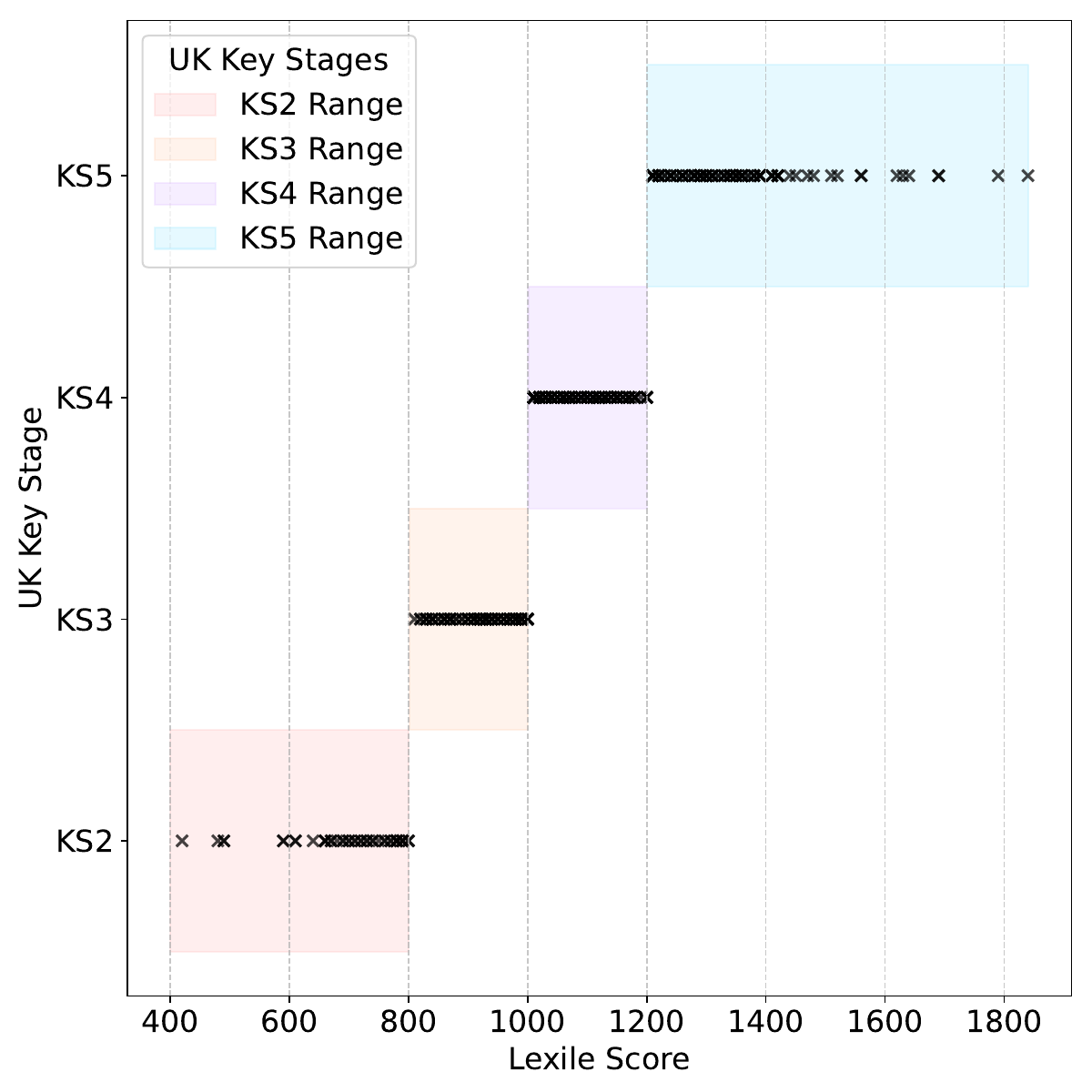}
        \caption{Distribution of Lexile Scores within the dataset.}
        \label{fig:scatter}
    \end{subfigure}
    \hfill
    \begin{subfigure}[b]{0.48\textwidth}
        \centering
        \includegraphics[width=\textwidth]{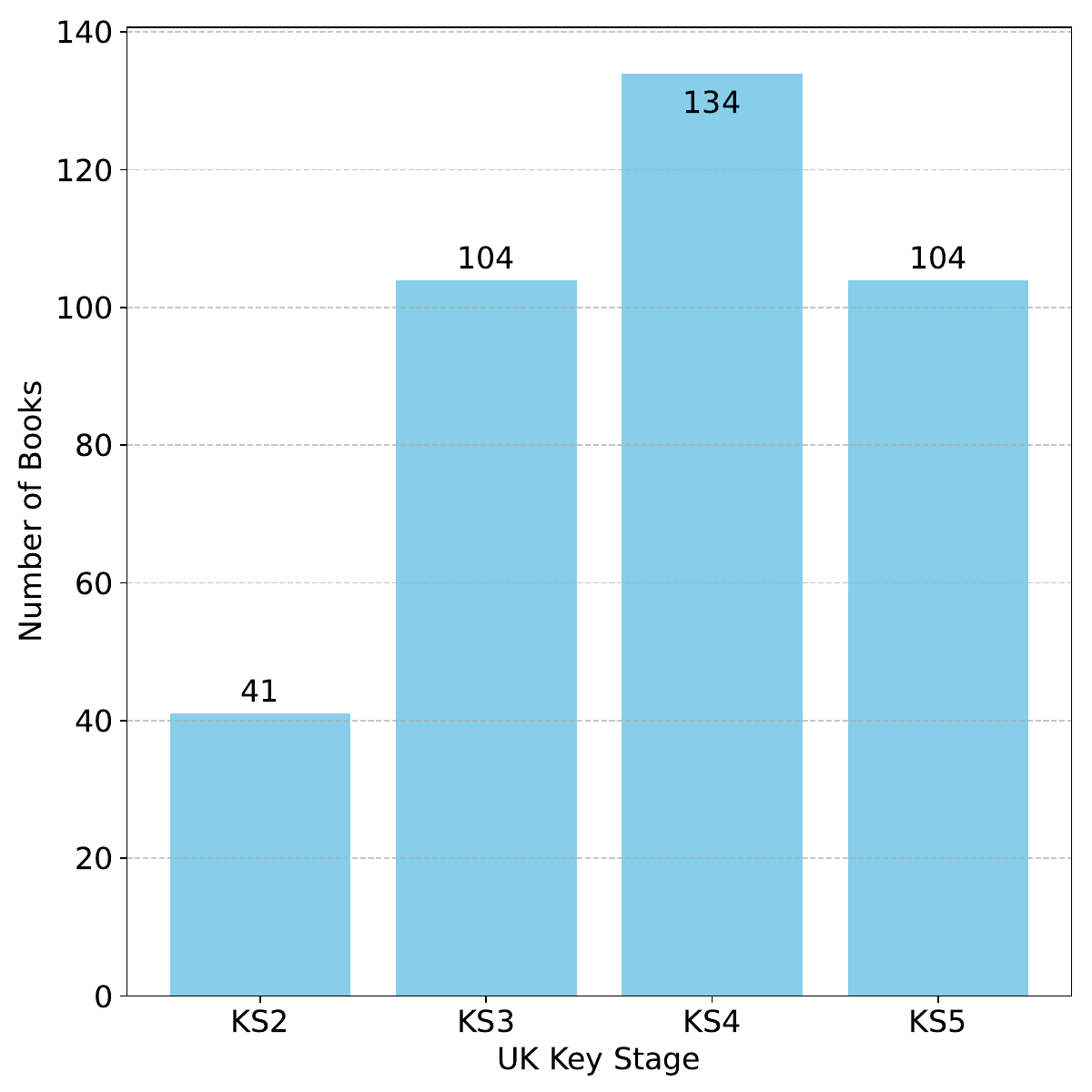}
        \caption{Total books belonging to each UK Key Stage.}
        \label{fig:bar}
    \end{subfigure}
    \caption{Overview of the pre-balanced dataset given Lexile scores and UK Key Stage categorisation.}
    \label{fig:distributions}
\end{figure}

Initial data collection was performed via Project Gutenberg, which is an online platform that distributes books within the public domain, i.e., those that can be used for non-commercial use without permission. Table \ref{tab:bookshelves} describes the repositories used for data collection. Each book that was available in raw text format was downloaded, leading to an initial set of 2009 books for further processing. Following this, the set of books was cross-referenced with the Lexile book finder. If a Lexile score was not available, the book was discarded from the dataset. Following this filter, a total of 384 books with Lexile scores were recovered and then the Lexile score was converted from a numerical value to a nominal class label according to Table \ref{tab:lexile}. A visualisation of the dataset prior to balancing can be observed in Figure \ref{fig:distributions}. Books belonging to Key Stage 1 were not available, and so are not considered in this study. The lowest score was \textit{The Monkey's Paw} by W.W. Jacobs (Oxford Bookworms) at 420, and the highest was \textit{Discourse on the Method of Rightly Conducting the Reason} by René Descartes at 1840. 

Given the data requirements of several state-of-the-art transformer models having a maximum input length of 512 tokens, each book was then divided into chunks of a maximum of 512 tokens to the nearest complete sentence. This resulted in a large and unbalanced dataset with a total of 515,688 rows. To alleviate issues of data size and balance, the dataset was then resampled by selecting 5000 rows per class. This resulted in a full dataset size of 20,000 data objects, with 5,000 belonging to each Key Stage 2, 3, 4, and 5. Finally, the dataset was split for training and validation at an 80/20 split, resulting in 4000 rows for training and 1000 rows for testing for each Key Stage. The data produced and utilised within this study is publicly available under the MIT license\footnote{Dataset available from:\\ \url{https://www.kaggle.com/datasets/birdy654/uk-key-stage-readability-for-english-texts}}.

\subsubsection{Linguistic Feature Generation}
From each of the text excerpts divided to the maximum length by the nearest complete sentence, the numerical linguistic characteristics were extracted within ten categories. The features were selected based on the criteria for producing fixed-length vectors, given that machine learning models require fixed input types for training. The features described in this section were extracted using the TextBlob \cite{loria2018textblob}, NRCLex \cite{mohammad2013crowdsourcing}, and NLTK \cite{loper2002nltk} Python libraries. The numerical features described in this Section were utilised for training the neural networks described in Section \ref{subsec:ml}. These features were calculated as follows:

\textbf{Basic Text Metrics} which include the number of words, sentences, unique words, and the average length of both sentences and words. 

\textbf{Detailed Sentence Information}, which includes the average number of characters and syllables per word, as well as the per-sentence averages of characters, syllables, words, types of words, paragraphs, long words, complex words, and Dale-Chall complex words. 

\textbf{Lexical Diversity and Richness} features, where \textit{diversity} refers to the variety of unique words within a text, and \textit{richness} refers to the sophistication of vocabulary within a text. The measures of lexical diversity and richness include the Type Token Ratio $TTR = \frac{V}{N}$, where $V$ is the number of unique words and $N$ is the total number of words. Higher values of $TTR$ suggest a greater variety in vocabulary. 

Yule's $K$: 
\begin{equation}
    K = 10^4 \times \frac{\sum_{i=1}^{V} i^2 \cdot f_i - N}{N^2},
\end{equation}
where $K$ is a quantification of richness, and $f_i$ is the frequency of the $i^{th}$ word type. Higher values suggest greater diversity in the text. 

Simpson's $D$: 
\begin{equation}
    D = \frac{\sum_{i=1}^{V} f_i(f_i - 1)}{N(N - 1)},
\end{equation}
where $D$ is the probability that two tokens selected at random are the same type, thus $D$ aims to measure the repeated use of words. A lower value of $D$ suggests that there is a higher diversity, since tokens are more likely to differ from one another.

Herdan's $C$: 
\begin{equation}
    C = \frac{\log N}{\log V},
\end{equation}
where $C$ is calculated by total words $N$ and unique words $V$. Lower values denote greater diversity since $V$ is relatively in comparison to $N$. 

Brunét's $W$: 
\begin{equation}
     W = N^{\left( V^{-0.165} \right)},
\end{equation}
for total words $N$ and unique words $V$, with a constant used to prevent distortions when presented with longer text sequences. Lower values of $W$ indicate a higher richness of vocabulary. 

Honoré's $R$: 
\begin{equation}
    R = 100 \times \frac{\log N}{1 - \frac{V_1}{V}}.
\end{equation}
$R$ is the relationship between total words $N$, unique words $V$, and words that appear only once (\textit{hapax legomena}) $V_1$. Higher values suggest rich vocabularies, especially when a wide range of infrequent words is used.

\textbf{Readability Scores} which estimate how difficult a text is to read, often related to the US educational grade levels. 

Kincaid Grade Level, which estimates the US educational grade level required to comprehend a given text: 
\begin{equation}
    \text{Kincaid} = 0.39 \left( \frac{\text{Total Words}}{\text{Total Sentences}} \right) + 11.8 \left( \frac{\text{Total Syllables}}{\text{Total Words}} \right) - 15.59.
\end{equation}

The Automated Readability Index (ARI). Similarly to the Kincaid level, ARI estimates the US grade level required to understand a text in relation to the number of characters: 
\begin{equation}
    \text{ARI} = 4.71 \left( \frac{\text{Characters}}{\text{Words}} \right) + 0.5 \left( \frac{\text{Words}}{\text{Sentences}} \right) - 21.43.
\end{equation}

Coleman-Liau Index, which is a prediction of the US grade level required for text comprehension given average characters per 100 words $L$ and average number of sentences per 100 words $S$:
\begin{equation}
    \text{Coleman-Liau} = 0.0588L - 0.296S - 15.8,
\end{equation}

The Flesch Reading Ease, which indicates the readability of a text given the observed lengths of words and sentences: 
\begin{equation}
    \text{Flesch} = 206.835 - 1.015 \left( \frac{\text{Total Words}}{\text{Total Sentences}} \right) - 84.6 \left( \frac{\text{Total Syllables}}{\text{Total Words}} \right).
\end{equation}

The Gunning Fog Index which estimates how many years of formal education would be required to understand a text on the first read: 
\begin{equation}
    \text{Gunning Fog} = 0.4 \left[ \left( \frac{\text{Words}}{\text{Sentences}} \right) + 100 \left( \frac{\text{Complex Words}}{\text{Words}} \right) \right].
\end{equation}

Läsbarhets Index (LIX) which is a score on how difficult a text is to read in relation to the lengths of words and sentences: 
\begin{equation}
    \text{LIX} = \frac{\text{Words}}{\text{Sentences}} + \frac{100 \times \text{Long Words}}{\text{Words}}.
\end{equation}

SMOG Index, which is an estimate of how many years of education are required to understand a text in relation to how many words are polysyllabic:
\begin{equation}
    \text{SMOG} = 1.0430 \sqrt{\text{Polysyllable Words} \times \frac{30}{\text{Sentences}}} + 3.1291.
\end{equation}

Andersson's Readability Index (RIX), which is a readability score in relation to how common long words are in the text:
\begin{equation}
    \text{RIX} = \frac{\text{Long Words}}{\text{Sentences}}
\end{equation}

The Dale-Chall Readability Formula, which scores readability based on words that US 4$^{th}$ grade students were observed to easily understand: 
\begin{equation}
    \text{Dale-Chall} = 0.1579 \left( \frac{100 \times \text{Difficult Words}}{\text{Words}} \right) + 0.0496 \left( \frac{\text{Words}}{\text{Sentences}} \right)
\end{equation}

\textbf{Sentence Structure}, which includes the count of part-of-speech tags within the given text. These tags include past principle verbs (VBN), 3$^{rd}$ person singular present tense verbs (VBZ), past-tense verbs (VBD), auxiliary verbs (VB or VBP), nominalisation (NN), and present participle verbs (VBG). Additionally, the mean $TTR$ of sentences $\frac{V}{N}$, the mean words per sentence, and the mean words per paragraph.

\textbf{Word Usage and Frequency} which includes the frequency of pronounds, function words, conjunction words, pronounds, and prepositions.

\textbf{Punctuation and Style} which includes the frequency of punctuation usage, sentences that begin with pronouns, interrogative words, articles, subordinations, conjuctions, or propositions. 

\textbf{Sentiment and Emotion} which includes the polarity of the sentiment of the text on a scale of $-1$ negative to $1$ positive, and the subjectivity of said senitmental value on a scale of $0$ to $1$ for a more opinionated sentiment. Individual emotion scores are also given for the detection of fear, anger, anticipation, trust, surprise, sadness, disgust, and joy. 

\textbf{Named Entity Recognition (NER)} which is the frequency of usage of PoS. The NERs counted include PERSON (an individual's name), NORP (Nationalities, Religious or Political Groups), FAC (Facilities/buildings), ORG (Organisations), GPE (Geo-Political Entities), LOC (Non-GPE locations), PRODUCT (manufactured objects), EVENT (named events), WORK\_OF\_ART (names of pieces of creativity), LAW (named legal works such as constitutions or acts), and LANGUAGE (names of natural languages).

\subsection{Machine Learning}
\label{subsec:ml}

\begin{figure}
    \centering
    \includegraphics[width=0.75\linewidth]{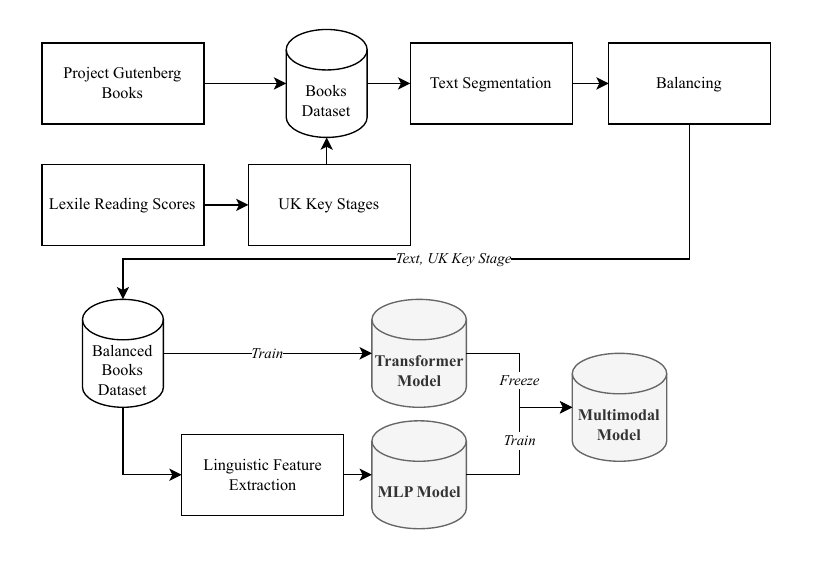}
    \caption{General diagram of the data generation and model training approaches followed in this study.}
    \label{fig:training-diagram}
\end{figure}

In this study, three types of approach were explored. First, for text classification, the use of Transformer models. Secondly, for the classification of numerical linguistic features, deep neural networks were explored. Finally, a fusion of the two was benchmarked through late fusion. This section describes the methods used for each of these approaches, respectively. A general overview of the proposed approach can be seen in Figure \ref{fig:training-diagram}.

\begin{table}[]
\centering
\caption{The models used in this study for text classification, sorted in descending order of size.}
\label{tab:transformers-list}
\footnotesize
\begin{tabular}{@{}lllll@{}}
\toprule
\textbf{Model} & \textbf{Parameters} & \textbf{Layers} & \textbf{Hidden Size} & \textbf{Attention Heads} \\ \midrule
Longformer\cite{beltagy2020longformer}     & 148M                & 12              & 768                  & 12                       \\
RoBERTa\cite{liu1907roberta}        & 125M                & 12              & 768                  & 12                       \\
XLNet\cite{yang2019xlnet}          & 117M                & 12              & 768                  & 12                       \\
ERNIE\cite{sun2019ernie}          & 109M                & 12              & 768                  & 12                       \\
BERT\cite{kenton2019bert}           & 109M                & 12              & 768                  & 12                       \\
ELECTRA\cite{clark2020electra}        & 110M                & 12              & 768                  & 12                       \\
DistilBERT\cite{sanh2019distilbert}     & 66M                 & 6               & 768                  & 12                       \\
ALBERT\cite{lan2019albert}         & 12M                 & 12              & 768                  & 12                       \\ \bottomrule
\end{tabular}
\end{table}

Toward the classification of the segmented text described in Section \ref{subsec:datacollection}, several transformer networks were selected for fine-tuning, which can be observed in Table \ref{tab:transformers-list}. A range of model sizes were selected from the current state of the art, from the largest model Longformer (148M parameters) down to the smallest model ALBERT (12M parameters). Each of the transformers were fine-tuned for 5 epochs on the text data. 

For the classification of the numerical linguistic features described in Section \ref{subsec:datacollection}, a random search of the hyperparameters of the artificial neural network was executed. 500 neural network architectures were searched of $\{16,...,256\}$ rectified linear units within $\{1,2,3,4,5\}$ hidden layers. The neural networks were trained with a learning rate of $0.001$ until the F1 score was not observed to increase in 15 epochs. 

In the final multimodality experiments, each Transformer was frozen, and the output layer was removed. Likewise, the deep neural network that performs best was selected with the output layer removed. Both models were fused by connecting them to a unified output layer, and the artificial neural network was trained on the linguistic features, with frozen transformer output also being used as input to the classification layer.

\subsection{Web Application for Inference and Reporting}
\begin{figure}
    \centering
    \includegraphics[width=0.75\linewidth]{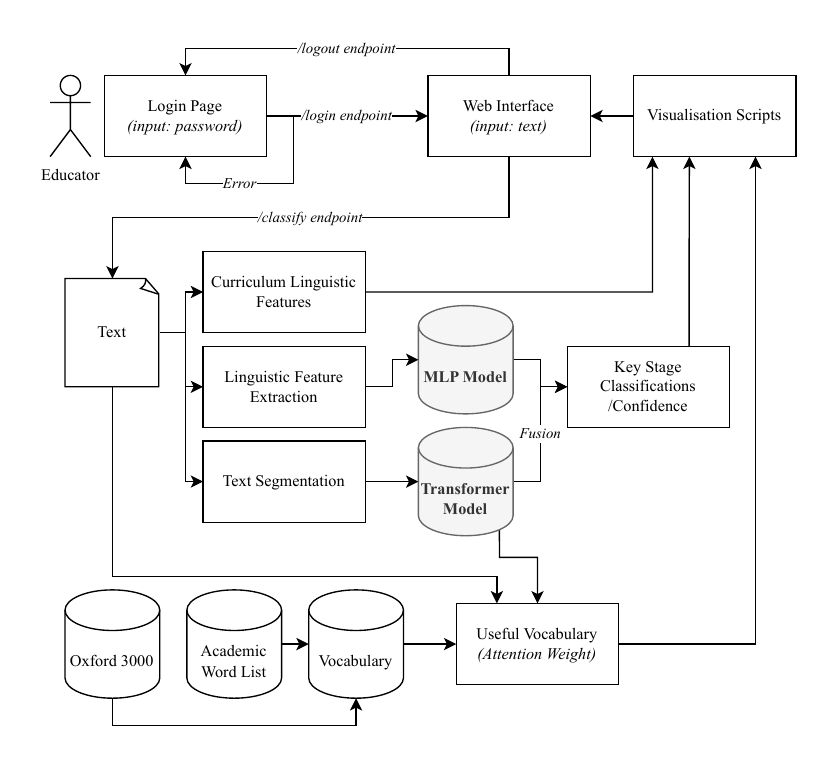}
    \caption{Flow Diagram for the web application which enables educators to utilise the machine learning and computational linguistics approaches.}
    \label{fig:webapp-flow}
\end{figure}

This subsection describes the method for integrating the models as well as NLP techniques within an interface designed to be used by non-technical stakeholders such as English teachers or librarians. The general approach to interface with the web application can be seen in Figure \ref{fig:webapp-flow}. Following authentication and input of a given text, the processes described in the previous sections are followed, and results are generated, which are passed back to the front-end application for asynchronous update and visualisation. The application provides a no-code interface for interaction with the linguistic feature extraction processes described in Subsection \ref{subsec:datacollection}, and the model inference process described in Subsection \ref{subsec:ml}. 

\begin{figure}
    \centering
    \includegraphics[width=0.4\linewidth]{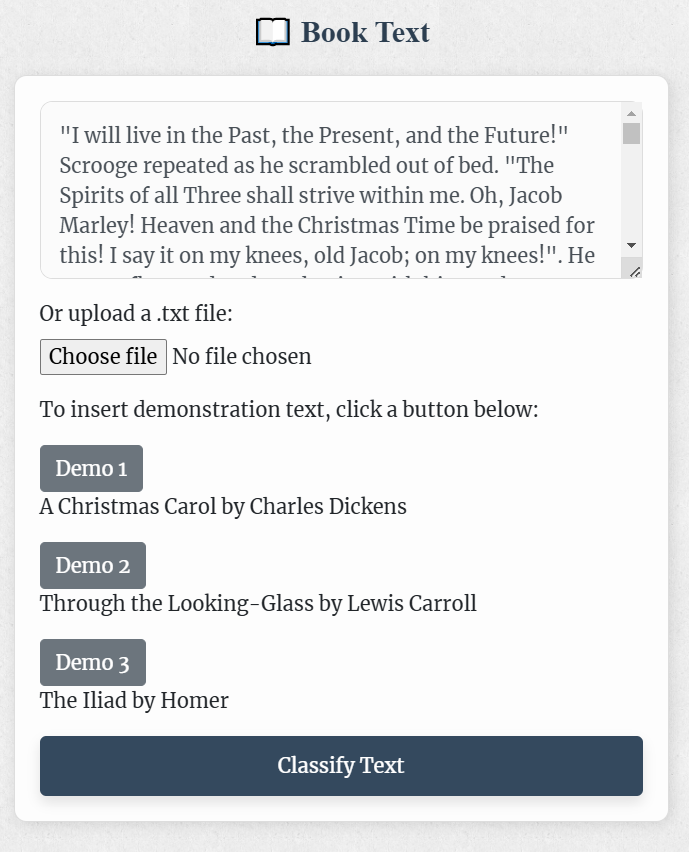}
    \caption{The container for educators to input text and run inference. Options include free text input, file upload, or demonstration excerpts.}
    \label{fig:text-input}
\end{figure}

Initially, educators input text into the system within the \textit{Book Text} container, as seen in Figure \ref{fig:text-input}. A free text box is available for typing or copy/paste, as well as file uploads. Additionally, demonstrations are available for A Christmas Carol by Charles Dickens, Through the Looking-Glass by Lewis Carroll, and Homer's Iliad. Upon clicking the \textit{demo} buttons, a stored excerpt is automatically entered into the input box. The \textit{Classify Text} button then communicates the text to the \textit{/classify} endpoint, and the web application is updated automatically through asynchronous processing when the classification is complete. 

Once the classification is complete, the following visualisations and information are provided to educators:

\begin{figure}
    \centering
    \includegraphics[width=0.6\linewidth]{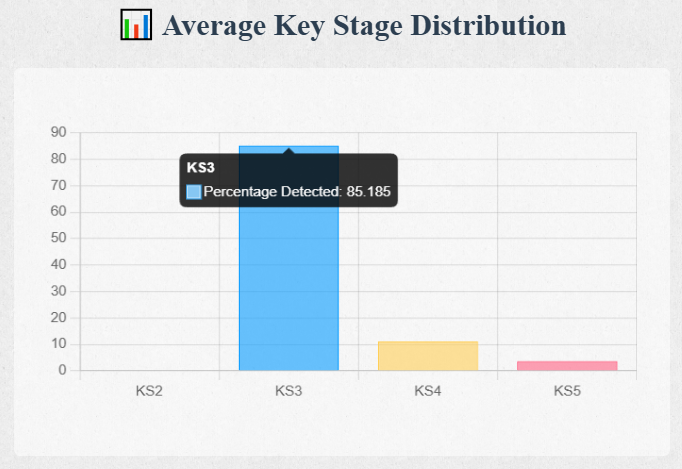}
    \caption{A visualisation provided to educators of the overall distribution of UK key stages detected in the provided text. Hovering the cursor over each bar provides a granular measure.}
    \label{fig:ks-distribution}
\end{figure}

First, the average distribution of each key stage detected in the text is visualised, as seen in Figure \ref{fig:ks-distribution}. Hovering the mouse cursor over each bar in the chart provides a more granular overview of how much of the text was predicted to be most appropriate for that key stage of education. The distribution is calculated by the percentage of text chunks that were predicted to belong to the key stage.

\begin{figure}
    \centering
    \includegraphics[width=0.7\linewidth]{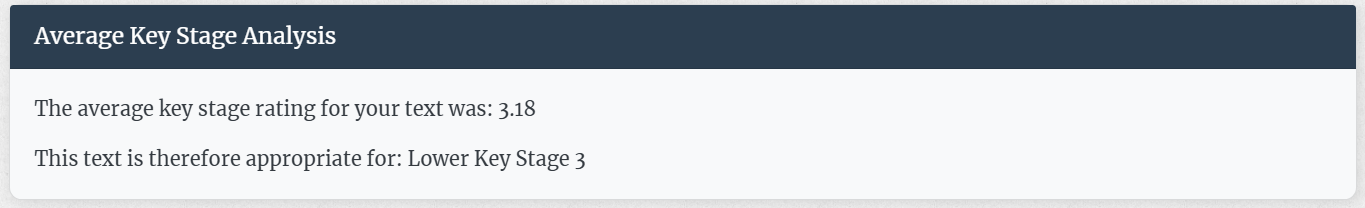}
    \caption{Information provided to the educators on the average key stage that was detected within the text, and a reading age recommendation is provided.}
    \label{fig:average-ks}
\end{figure}

\begin{figure}
    \centering
    \includegraphics[width=0.7\linewidth]{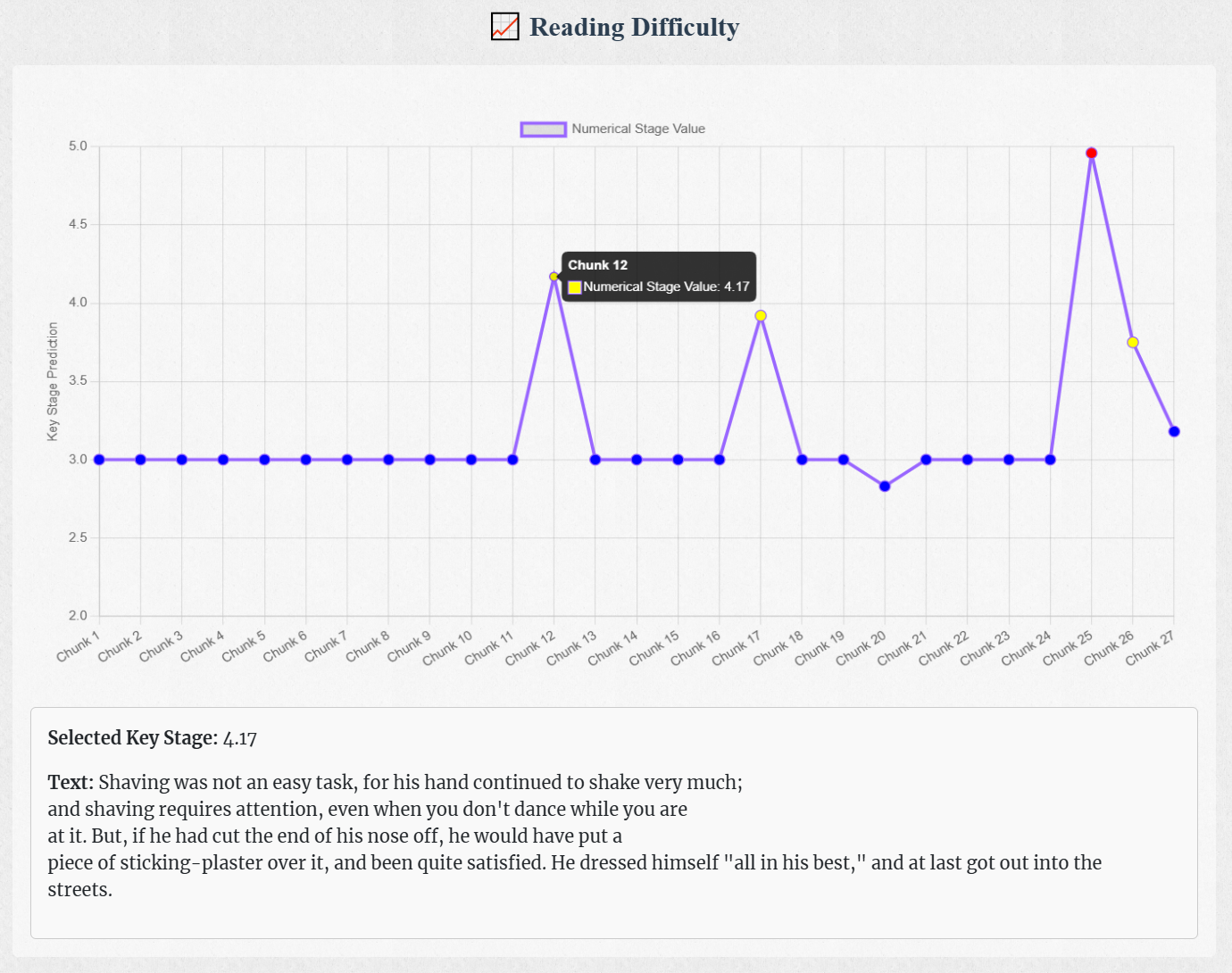}
    \caption{A visualisation provided to educators of the temporal (start-to-finish) predictions made on each chunk of the input text. Clicking on a point on the graph will provide the chunk of text.}
    \label{fig:reading-difficulty}
\end{figure}

The overall score is presented to the educators, as seen in Figure \ref{fig:average-ks}. The overall score is calculated by considering the Key Stage value $KS_j$ for each chunk $j$ (i.e. $KS_j \in \{2, 3, 4, 5\}$), and, for each prediction, the confidence score $C_j$, where $0 \leq C_j \leq 1$. The score is thus calculated as: 
\begin{equation}
    \label{eq:overallscore}
    \text{Overall Score} = \frac{\sum_{j=1}^{N} KS_j \cdot C_j}{\sum_{j=1}^{N} C_j}.
\end{equation}
Similarly, a more granular breakdown is provided within the \textit{reading difficulty} visualisation, an example of which can be observed in Figure \ref{fig:reading-difficulty}. Similarly to the overall score described in Equation \ref{eq:overallscore}, the individual score of a text chunk is calculated as $\sum_{i=2}^{5} KS_i \cdot P(KS_i)$. 

\begin{figure}
    \centering
    \includegraphics[width=0.6\linewidth]{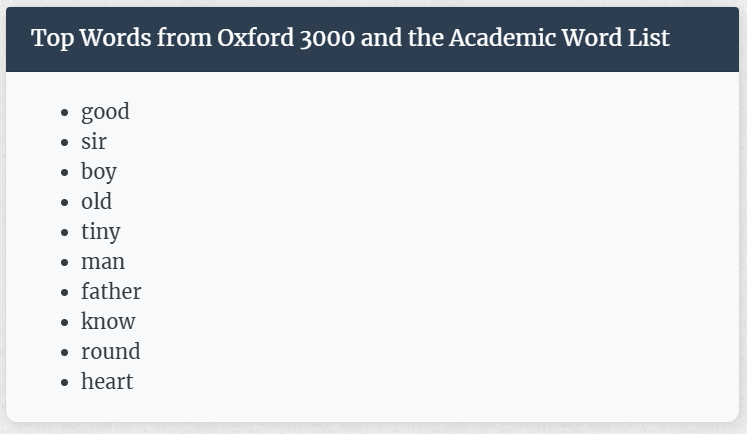}
    \caption{Information provided to educators of the top tokens used for classification that also exist within either the Oxford 3000 or Academic Word List lists.}
    \label{fig:top-words}
\end{figure}

The key vocabulary is selected from the text and presented to the educators as seen in Figure \ref{fig:top-words}. The word lists selected within this study are the Oxford 300 list\cite{oxford3000}, a curated list of the 3000 most important words for English learners, and the Academic Word List (AWL)\cite{coxhead2000new}, which is a collection of 570 word families commonly found in academic texts. For each of the tokens in the text $T = \{t_1,t_2,...,t_k\}$ that is contained within the aggregate lists (i.e., $t_j \in W$), the importance of each is calculated via the aggregated attention weight $A(t_j)$:
\begin{equation}
    A(t_j) = \frac{1}{L \cdot H} \sum_{l=1}^{L} \sum_{h=1}^{H} \sum_{i=1}^{N} A^l_{h, ij},
\end{equation}
for a transformer with $L$ layers, $H$ attention heads, $N$ tokens at index $i$. All tokens within the two lists are then sorted in descending order of aggregate attention, and the top ten are returned. The maximum of ten top tokens is an arbitrarily selected value which can be changed. 

\begin{figure}
    \centering
    \includegraphics[width=0.5\linewidth]{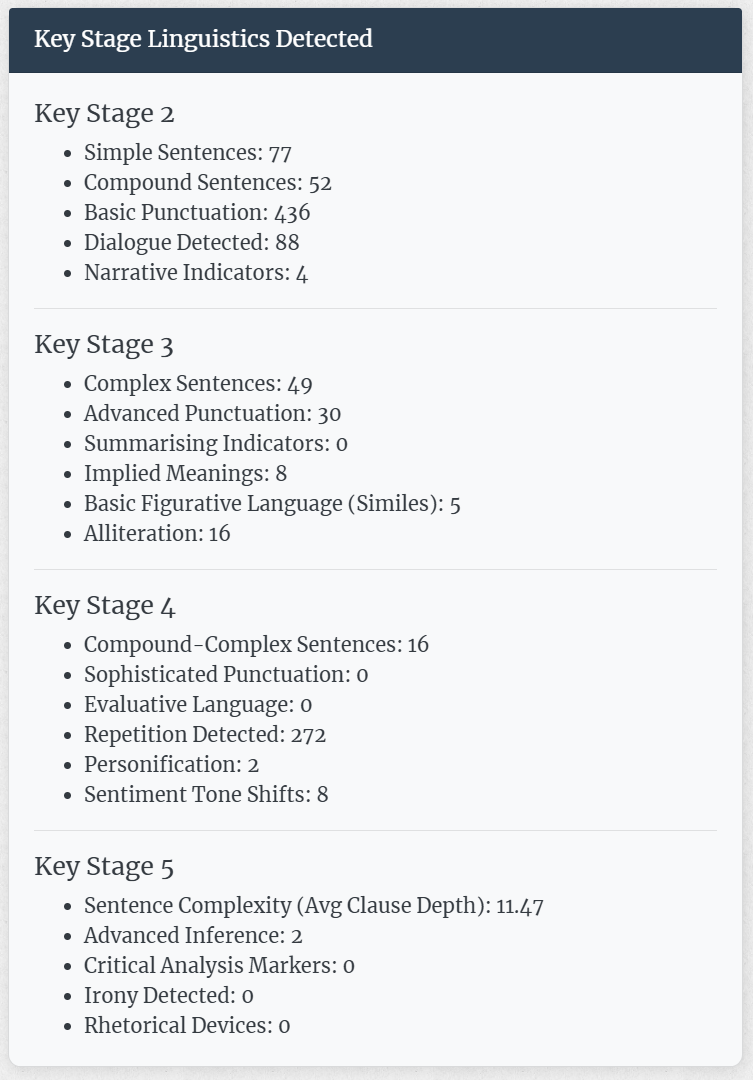}
    \caption{Information provided to educators of the linguistic features detected in the entire input text, categorised according to the National curriculum in England: English programmes of study.}
    \label{fig:linguistics-detected}
\end{figure}

The number of occurrences of multiple linguistic features are then detected and communicated from the entire text based on the statutory guidance \textit{National curriculum in England: English programmes of study}\cite{department2013national}. Detection is performed using the spaCy library\cite{spacy2} and regular expressions, and includes the following:

\textbf{For Key Stage 2:} Simple Sentences, which are defined as sentences with a dependency subtree containing ten or fewer words. Compound Sentences, which are defined as sentences that contain conjuctions such as \textit{and} or \textit{but}. Conjuctions are detected through the spaCy dependency tag \textit{cc}. Basic Punctuation through regular expression pattern matching for fullstops, commas, exclamation, and question marks. Dialogue, defined as sentences that contain quotation marks. Narrative Indicators common in storytelling such as \textit{then}, \textit{next}, \textit{afterwards}, etc.

\textbf{For Key Stage 3:} Complex Sentences are defined as sentences that contain subordinating conjuctions (that is, \textit{because}, \textit{although}, etc.) or adverbial clauses. These are detected using spaCy via the \textit{mark} and \textit{advcl} dependency tags. Advanced Punctuation through regular expression pattern matching for colons, semicolons, and parentheses. Summarizing Indicators through keyword matching of phrases such as \textit{in summary}, \textit{to conclude}, \textit{overall}, etc. Implied meanings through keyword matching of phrases that convey inferential or conditional reasoning such as \textit{if}, \textit{unless}, \textit{suggests}, etc. Figurative Language (Similes) defined as phrases that contain similies (\textit{like}, \textit{as}, etc.) with adjectival modifiers (e.g., \textit{eyes as deep as the ocean}). This dependency is detected using spaCy with relation tag \textit{amod}. Alliteration, where sentences contain repeated initial sounds used for stylistic effect.

\textbf{For Key Stage 4:} Compound-Complex Sentences which are defined as sentences that contain both coordinating and subordinating conjuctions, detected with the spaCy tags \textit{cc} and \textit{mark}. Sophisticated Punctuation through regular expression pattern matching for dashes and ellipses used in advanced sentence structuring. Evaluative Language through detecting terms that indicate judgement or assessment, such as \textit{valid}, \textit{effective}, etc. Repetition through detecting repeated words that indicate emphasis or redundancy. Personification through named entity recognition (\textit{PERSON} label) and dependency parsing to identify human-like actions. Tone Shifts through detecting shifts in argument or sentiment (e.g. \textit{however}, \textit{but}, \textit{nevertheless}, etc.)

\textbf{For Key Stage 5:} Advanced Inference through the detection of logical markers such as \textit{therefore}, \textit{hence}, etc. Critical Analysis through the detection of phrases such as \textit{persuasive}, \textit{flawed}, etc. which indicate a critical evaluation. Irony through the detection of phrases that contain contrastive terms and through dependency parsing within spaCy. Rhetorical Devices through the detection of structural patterns such as \textit{not only ... but also}.

\begin{figure}
    \centering
    \includegraphics[width=0.8\linewidth]{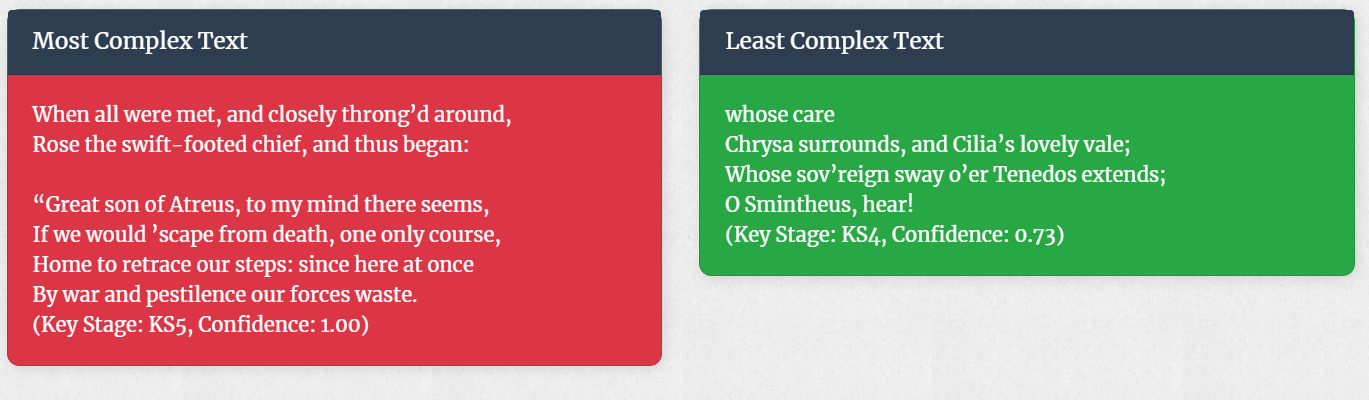}
    \caption{Information provided to educators on the most and least complex excerpts from the given text. Calculated via the highest key stage prediction with the highest confidence, and lowest key stage prediction with the highest confidence, respectively.}
    \label{fig:most-least-complex}
\end{figure}

Finally, the most and least complex excerpts of the given text are defined as the highest predicted key stage with the highest confidence and the lowest predicted key stage with the highest confidence, respectively. An example of these excerpts can be found in Figure \ref{fig:most-least-complex}.

\subsection{Experimental Setup}
The experiments in this work were executed on a server containing 4 Nvidia RTX A2000 GPUs. Each GPU had 3,328 CUDA cores and 12GB of VRAM. For training, all four GPUs were used. For inference, one GPU was used in terms of processing, but it was observed that temporary model storage on the VRAM was shared across each of the four units. All models were trained via the PyTorch library\cite{NEURIPS2019_9015}, and the educator-facing web application was developed using the Flask framework\cite{grinberg2018flask}.

\section{Results and Discussion}
\label{sec:resultsdiscussion}

\begin{table}[]
\centering
\caption{Overall results of all models sorted by F1 score.}
\label{tab:all-results}
\footnotesize
\begin{tabular}{@{}lllllll@{}}
\toprule
\textbf{Model}   & \textbf{Accuracy} & \textbf{Precision} & \textbf{Recall} & \textbf{F1} & \textbf{Parameters} & \textbf{\begin{tabular}[c]{@{}l@{}}Inference \\ Time (S)\end{tabular}} \\ \midrule
ELECTRA + ANN    & 0.997             & 0.997              & 0.997           & 0.996       & 108907499           & 0.018                                                                  \\
ERNIE + ANN      & 0.995             & 0.995              & 0.995           & 0.994       & 109499627           & 0.018                                                                  \\
XLNet + ANN      & 0.992             & 0.992              & 0.992           & 0.992       & 116734187           & 0.025                                                                  \\
RoBERTa + ANN    & 0.987             & 0.988              & 0.987           & 0.987       & 124661483           & 0.019                                                                  \\
DistilBERT + ANN & 0.987             & 0.987              & 0.987           & 0.987       & 66378731            & 0.011                                                                  \\
Longformer + ANN & 0.939             & 0.951              & 0.939           & 0.939       & 148675307           & 0.040                                                                  \\
BERT + ANN       & 0.905             & 0.905              & 0.905           & 0.905       & 109498091           & 0.018                                                                  \\
ALBERT + ANN     & 0.741             & 0.862              & 0.741           & 0.797       & 11699435            & 0.010
\\
BERT             & 0.750             & 0.751              & 0.750           & 0.750       & 109485316           & 0.010                                                                  \\
DistilBERT       & 0.744             & 0.744              & 0.744           & 0.744       & 66956548            & 0.006                                                                  \\
Longformer       & 0.741             & 0.741              & 0.741           & 0.740       & 148662532           & 0.036                                                                  \\
XLNet            & 0.742             & 0.740              & 0.742           & 0.740       & 117312004           & 0.022                                                                  \\
ERNIE            & 0.735             & 0.740              & 0.735           & 0.736       & 109486852           & 0.011                                                                  \\
RoBERTa          & 0.731             & 0.731              & 0.731           & 0.731       & 124648708           & 0.010                                                                  \\
ELECTRA          & 0.714             & 0.713              & 0.714           & 0.713       & 109485316           & 0.011                                                                  \\
ALBERT           & 0.675             & 0.685              & 0.675           & 0.678       & 11686660            & 0.009                                                                                                                                  \\ \bottomrule
\end{tabular}
\end{table}

This section contains the results of the experiements described previously. The results of all the experiments can be found in Table \ref{tab:all-results}. The best single modal model overall was BERT, which had an F1 score of 0.75. It can be observed that all multimodal models outperformed all transformer-based unimodal text classifiers. The overall best model was the multimodal combination of the ELECTRA text classifier and linguistics artificial neural network, which achieved an F1 score of 0.996.

\begin{figure}[]
    \centering
    \begin{subfigure}[b]{0.45\textwidth}
        \centering
        \includegraphics[width=\textwidth]{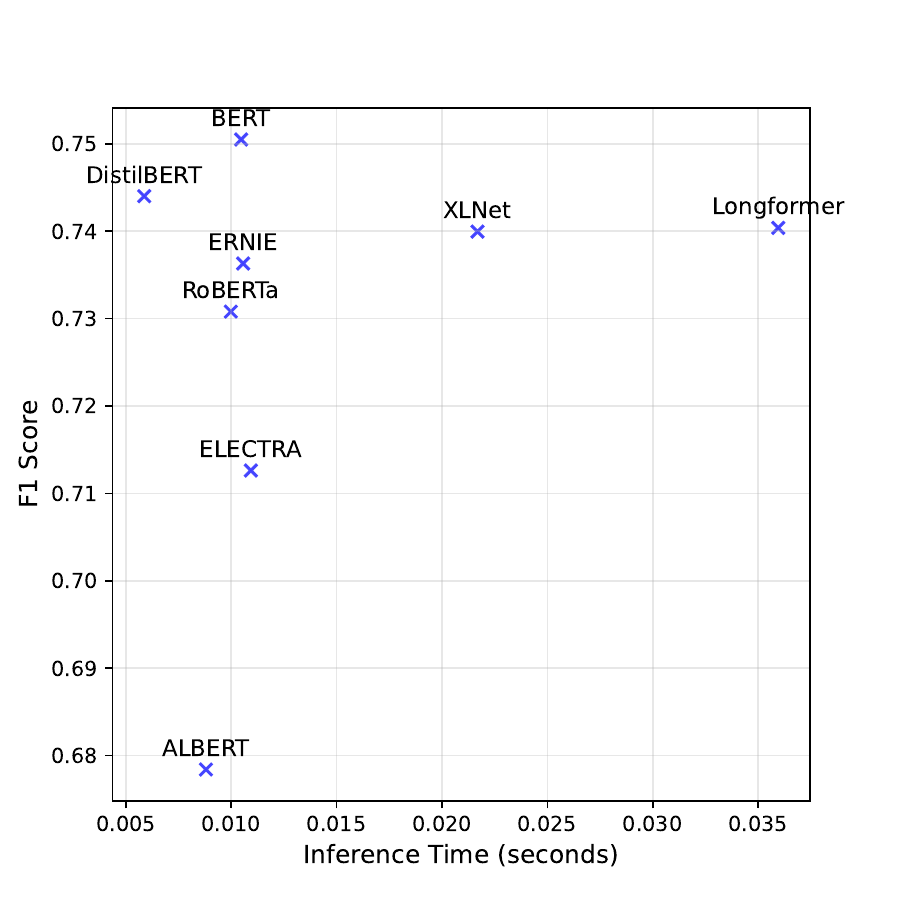}
        \caption{Transformer text classifiers.}
        \label{fig:unimodalplot-tf}
    \end{subfigure}
    \hfill
    \begin{subfigure}[b]{0.45\textwidth}
        \centering
        \includegraphics[width=\textwidth]{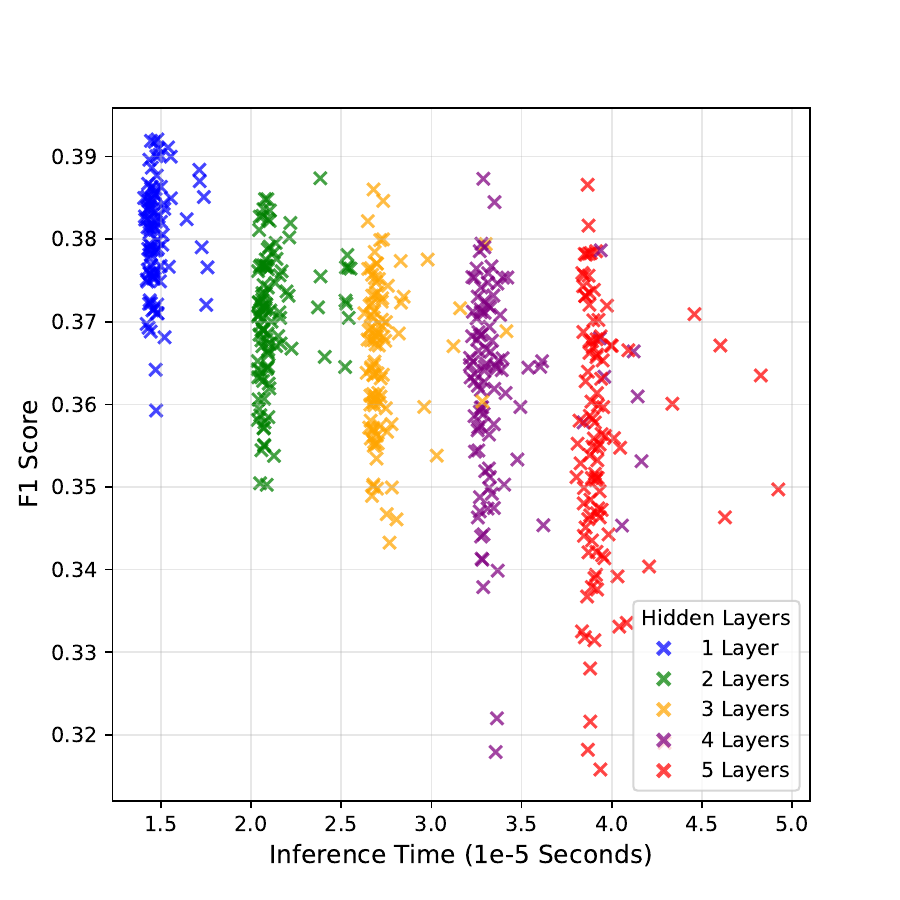}
        \caption{ANN linguistics classifiers.}
        \label{fig:annplot}
    \end{subfigure}
    \caption{Comparison of the results for both types of unimodal model.}
    \label{fig:unimodals}
\end{figure}

Figure \ref{fig:unimodals} shows the best results for the unimodal classifiers. The best text classifier was BERT, with an F1 score of 0.75. It can be observed that neural networks trained on linguistic features perform poorly, with a pattern of higher network depth that leads to worse overall results. The best neural network had one hidden layer of 175 rectified linear units, which scored 40.2\% accuracy, 0.394 precision, 0.402 recall, and an F1 score of 0.392. The neural network had 12604 trainable parameters in total and could infer data objects at an average speed of 1.48 1e-05 seconds. The network containing 1 layer of 175 hidden neurons is carried forward to the multimodal experiments, while the others are discarded henceforth. 

\begin{figure}
    \centering
    \includegraphics[width=0.8\linewidth]{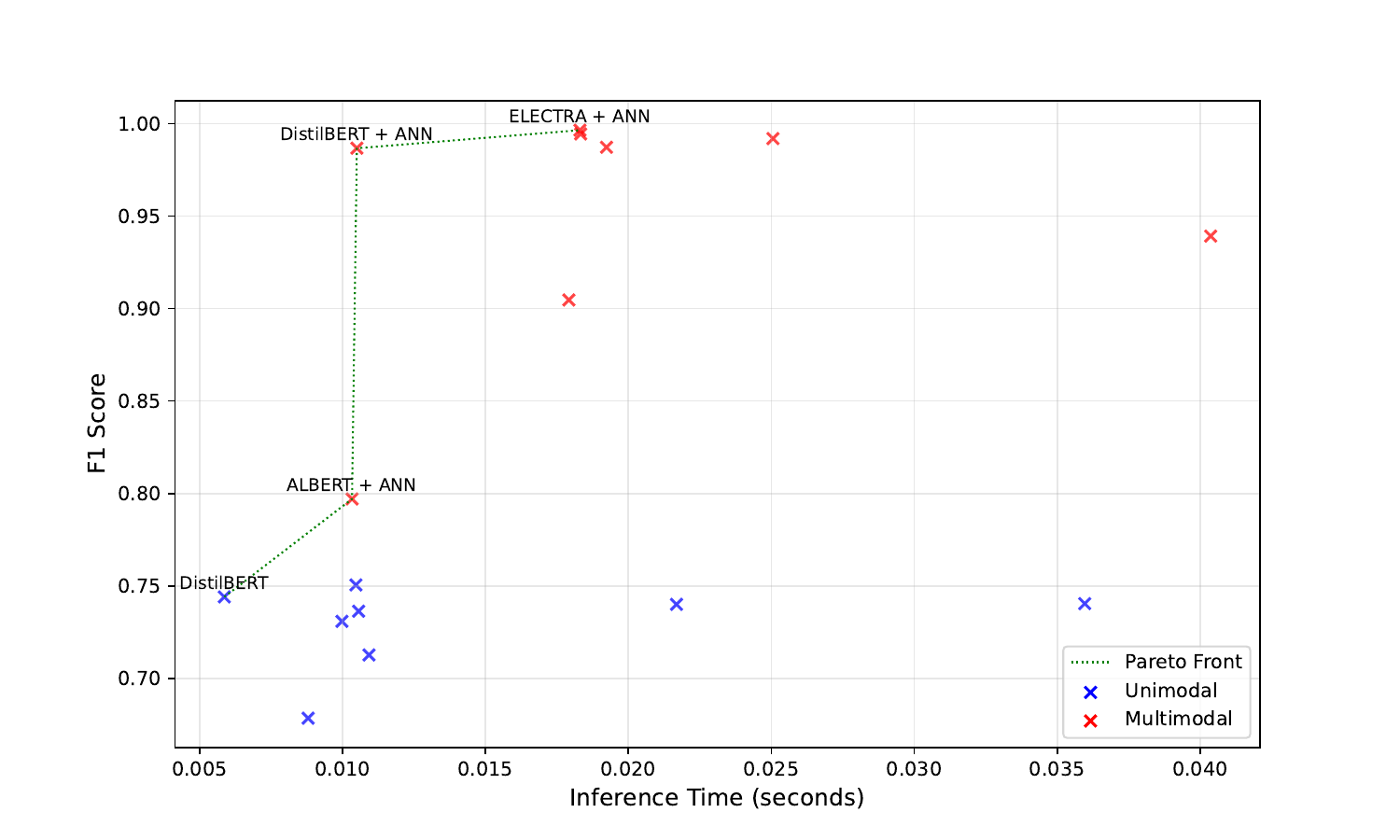}
    \caption{Pareto optimality graph of all single and multimodal models for F1 score and inference time.}
    \label{fig:pareto}
\end{figure}

The results of the pareto optimality can be observed in Figure \ref{fig:pareto}. In terms of a higher F1 score and a shorter inference time, the four models that were not completely dominated by any other model were the unimodal DistilBERT, and the multimodal ALBERT, DistilBERT, and ELECTRA models. 
\begin{figure}[]
    \centering
    \begin{subfigure}[b]{0.4\textwidth}
        \centering
        \includegraphics[width=\textwidth]{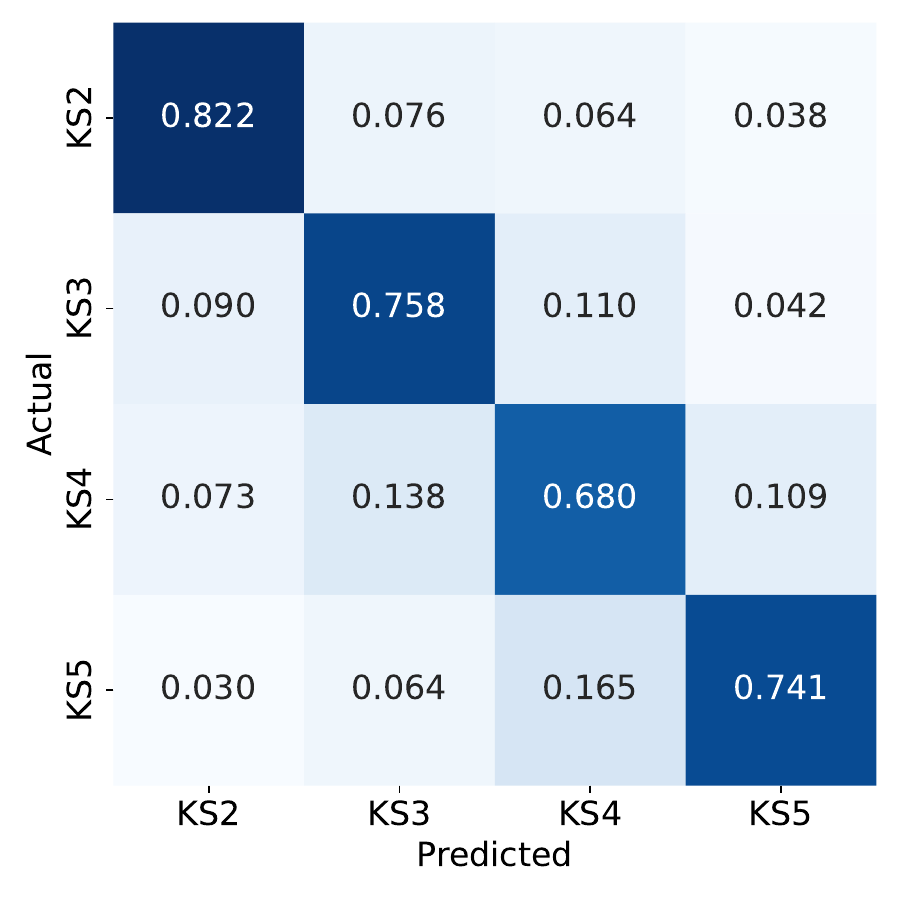}
        \caption{BERT (Unimodal)}
        \label{fig:confusion1}
    \end{subfigure}
    \hspace{1.5cm}
    \begin{subfigure}[b]{0.4\textwidth}
        \centering
        \includegraphics[width=\textwidth]{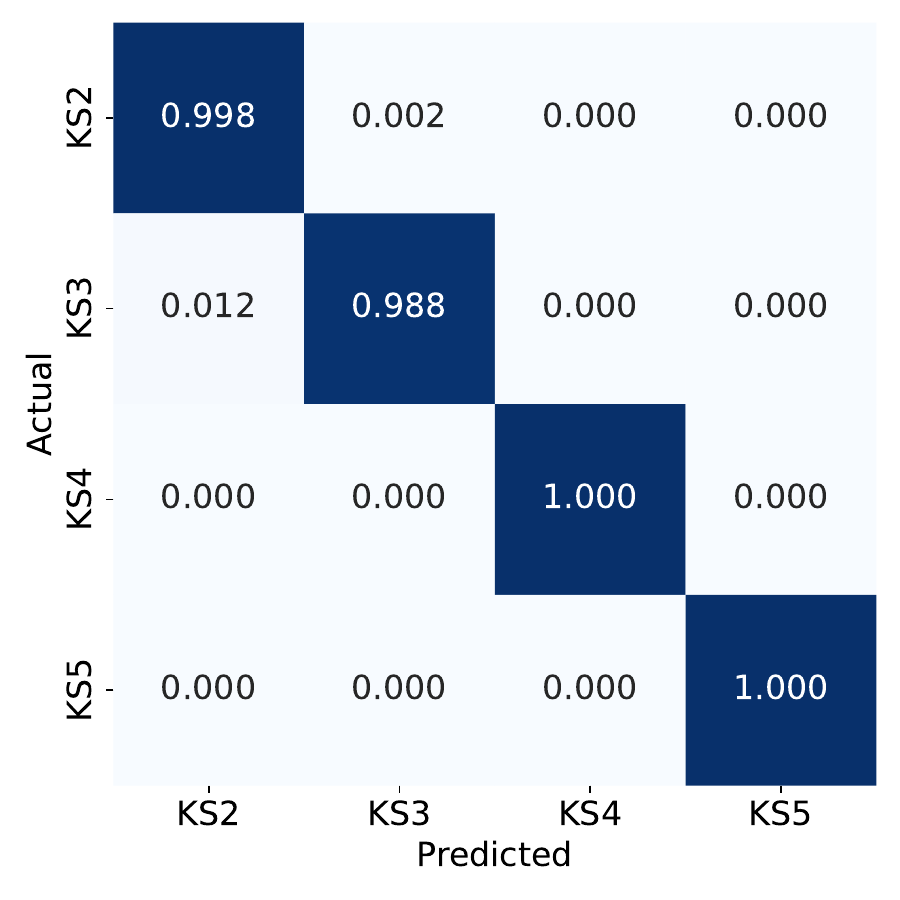}
        \caption{ELECTRA (Multimodal)}
        \label{fig:confusion2}
    \end{subfigure}
    \caption{Comparison of the normalised confusion matrices of the best unimodal and multimodal approaches.}
    \label{fig:confusion_matrices}
\end{figure}
The confusion matrices for the best performing unimodal (BERT) and multimodal (ELECTRA) approaches can be found in Figure \ref{fig:confusion_matrices}. The most problematic class for the unimodal approach was Key Stage 4, with several excerpts misinterpreted as Key Stage 3. However, the best-performing multimodal approach was able to classify Key Stages 4 and 5 perfectly, with relatively lower confusion between Key Stages 2 and 3. 

\begin{table}[]
\caption{Results of the paired t-test for all metrics when comparing unimodal to multimodal approaches. }
\label{tab:stat-tests}
\centering
\begin{tabular}{@{}llll@{}}
\toprule
\textbf{Metric}    & \textbf{t-statistic} & \textbf{p-value} & \textbf{p \textless 0.05} \\ \midrule
Accuracy           & 9.45                 & \textless{}0.001 & Yes                       \\
Precision          & 21.95                & \textless{}0.001 & Yes                       \\
Recall             & 9.45                 & \textless{}0.001 & Yes                       \\
F1                 & 13.30                & \textless{}0.001 & Yes                       \\
Inference Time (S) & 1.27                 & 0.244            & No                        \\ \bottomrule
\end{tabular}
\end{table}
Following the observation that all multimodal models seemed to outperform unimodal approaches, further validation is performed through paired t-tests on the performance metrics of accuracy, precision, recall, F1 score and inference time. The results can be observed in Table \ref{tab:stat-tests}, which shows that these improvements in the classification metrics are statistically significant ($p<0.05$). Furthermore, the paired t-test for inference time ($p = 0.244$) shows that there is no statistically significant difference, suggesting that the addition of the artificial neural network does not have a considerable impact on processing time.

\section{Conclusion and Future Work}
\label{sec:conclusion}
This study has explored the use of multimodal fusion for the classification of educational literature into appropriate UK Key Stages. Results showed that multimodal approaches significantly outperformed their constituent unimodal models. The proposed framework and web application combines linguistic feature analysis with transformer-based text classification, and forms a robust tool for non-technical stakeholders to rapidly assess the complexity and suitability of different educational texts. Among the models evaluated, it was found that the multimodal ELECTRA and ANN approach achieved the highest classification ability with an F1 score of 0.996.

The application developed to deliver the work performed in this article enables educators to analyse text and retrieve granular insights into Key Stage alignment. This real-time, stakeholder-facing tool addresses open challenges in the field, allowing rapid response to emerging trends for encapsulation within the education system, as well as reducing manual workload for educators. Future work could focus on expansion beyond the current data towards internationalisation, that is, tuning the models for appropriate in education systems outside the UK. Given the findings of this study, future work could also consider user feedback from domain experts such as teachers or librarians, with the aim of greater real-world impact. In terms of the methodology of this work, although multimodal models outperformed all unimodal models, the unimodal linguistics models were significantly weaker. Given this scientific limitation, further work could be explored in improving linguistic models prior to multimodal fusion. Due to limitations of computational resources, this study undersampled the full dataset, and thus future work could consider benchmarking the approaches on the full set of texts. 

By alleviating some of the open challenges in the field of educational text analysis, this work has proposed approaches towards scalable, data-driven tools that have the potential to empower educators to adapt teaching materials dynamically with little manual effort required. 

\section{Data Availability Statement}
The data produced and utilised within this study is publicly available under the MIT license\footnote{Dataset available from:\\ \url{https://www.kaggle.com/datasets/birdy654/uk-key-stage-readability-for-english-texts}}.

\bibliographystyle{ieeetr}
\bibliography{bibliography}

\end{document}